# ESG-FTSE: A corpus of news articles with ESG relevance labels and use cases

**Mariya Pavlova. Miaosen Wang, Bernard Casey**

Imperial College London, Google DeepMind, SOCial ECONomic RESearch London and Frankfurt
Exhibition Rd, South Kensington, London SW7 2BX, United Kingdom
1600 Amphitheatre Parkway, Mountain View, CA, United States
16 Sandringham Court, Arregon Road, London TW1 3ND, United Kingdom
m.pavlova22@imperial.ac.uk, miaosen@deepmind.com, b.casey@soceconres.eu

**Abstract**

We present ESG-FTSE, the first corpus comprised of news articles with Environmental, Social and Governance (ESG) relevance annotations. In recent years, investors and regulators have pushed ESG investing to the mainstream due to the urgency of climate change. This has led to the rise of ESG scores to evaluate an investment's credentials as socially responsible. While demand for ESG scores is high, their quality varies wildly. Quantitative techniques can be applied to improve ESG scores, thus, responsible investing. To contribute to resource building for ESG and financial text mining, we pioneer the ESG-FTSE corpus. We further present the first of its kind ESG annotation schema. It has three levels: a binary classification (relevant versus irrelevant news articles), ESG classification (ESG-related news articles), and target company. Both supervised and unsupervised learning experiments for ESG relevance detection were conducted to demonstrate that the corpus can be used in different settings to derive accurate ESG predictions.

## 1. Introduction

ESG is a framework that aims to capture all the non-financial information arising from a company's day-to-day activities. Financial markets have been going through a seismic shift with the rise of ESG investing. The pressing need to address climate change has led to the ascent of sustainable investing. This has also boosted the proliferation of ESG scores. Their different requirements and quality have added cost, confusion, risk, and complexity to investors. According to recent research, poor data quality is one of the biggest obstacles in ESG investing (Murray, 2021). This, in turn, has prompted concerns over "greenwashing" – i.e., that some investments are not as sustainable as they claim to be. Subsequently, this has negatively impacted the fight against global warming. In 2023, the European Commission sought responses from relevant regulators and has pressed for better disclosure (Jones, 2023). ESG scores are the most widely used metric. Yet, they have been scrutinised by regulators and investors because of their questionable quality. We argue that the limitations of the research methods used to generate ESG scores are one of the main barriers to responsible investing. We discuss this in section 2.1.

Artificial Intelligence (AI) techniques can drastically improve the accuracy of ESG scores and, thus, ESG investing by automatically detecting socioeconomic events and news items that influence ESG scores. Despite that, research in this domain is limited. To encourage it, we take the novel and difficult approach of creating a corpus with ESG relevance labels. To the best of our knowledge, there is no such publicly available corpus. Annotating news articles with ESG context and relevance to a company poses several challenges. ESG relevance detection is highly dependent on one's ESG domain knowledge and own perspective on what constitutes ESG relevance to a company. The lack of a universal ESG score framework further complicates this. In addition, categorising ESG content is highly contextual. Given these challenges, we argue that providing a comprehensive annotation schema is crucial to ensuring consistency and good performance of natural language processing (NLP) tasks. To promote standardisation, our schema was inspired by the EU taxonomy for sustainable initiatives and the United Nations Sustainable Development Goals (SDGs).[1] [2]

Alternative data, such as news articles, are well-documented as a powerful tool for evaluating stock market performance and investment opportunities. By building a corpus that consists entirely of publicly available news articles about FTSE 100 Index constituents, we demonstrate that this approach can also be used for assessing ESG credentials.[3] Specialists with industry and academic experience in ESG, sustainable investing, economics, and finance manually annotated the ESG-FTSE corpus. Further, the annotators undertook training and followed precise guidelines to minimise bias and ensure the accuracy and consistency of the process. Baseline supervised and unsupervised learning experiments were conducted with the corpus. Results demonstrate high ESG salience and that NLP techniques can successfully be applied to the corpus for sustainability research. In addition, our experiments reveal that ESG has small data characteristics, i.e., scarce but relevant data. Despite that, experiments prove that it is feasible to make accurate ESG predictions even from small-volume data. The main contributions of this paper are the following:

- The first corpus with ESG relevance labels: the ESG-FTSE corpus. It consists of 3,913 news articles in English covering the period from late 2018 to summer 2021. To ensure the corpus is suitable for analysing a company's credentials, the news pieces are about the top ten FTSE 100 Index constituents by market capitalisation. By building the corpus entirely from publicly available news articles, we take the view that such alternative data influence not just financial performance but also a company's credentials, thus making it essential to ESG analysis.

- The first of its kind ESG annotation methodology. It consists of a three-level schema: a binary classification (relevant versus irrelevant news articles), ESG classification (environmental, social and governance-related news articles), and target company.

- We revealed a small data characteristic associated with ESG data, i.e. scarce but relevant data. This is an important characteristic of the ESG domain. Acute events, such as Covid-19 and extreme weather, are becoming more frequent. Thus, it is important to be able to utilise small-volume data to accurately predict such events. Our experiments prove that ESG-FTSE can be useful in making accurate ESG predictions.

---

[1] https://finance.ec.europa.eu/sustainable-finance/tools-and-standards/eu-taxonomy-sustainable-activities_en
[2] https://sdgs.un.org/goals
[3] https://www.londonstockexchange.com/indices/ftse-100

## 2. Related Work

### 2.1. Notion of ESG Investing

ESG investing is an umbrella term for investments that seek positive returns and long-term impact on society, the environment, and the business. Environmental criteria may consider an organisation's pollution, waste, energy use, natural resource conservation, carbon footprint, and treatment of animals. The case of the miner BHP damaging aboriginal sites, which prompted an inquiry in the Australian parliament, is an example. Social criteria examine a company's management and its relationships with employees, customers, suppliers, and the communities where it operates. The case of wages and conditions of workers in the Leicester garment factory, which led to retailers reconsidering their purchasing policies, is an example. Governance looks at a company's leadership, executive pay, audits, internal controls, lawsuits, and shareholder rights. The case of increases paid to AstraZeneca investors, which sparked a rejection by shareholders, is an example. Previously, ESG investing represented a niche area of financial markets. With regulators and investors realising the financial materiality of ESG risks, these financial products have experienced soaring demand. According to a report by Morgan Stanley, sustainable funds' assets under management (AUM) totalled nearly $2.8 trillion in 2022. They are continuing to grow as a proportion of overall AUM: 7% compared to 4% five years ago 2023. Their popularity has led to investors seeking more information on sustainability risks. This has given rise to various initiatives to define ESG disclosure standards, investment and measurement principles, and metrics (Murray, 2021). ESG scores have become the most widely used metric to measure a company's credentials. While they are high in demand, the same cannot be said for their quality. The multitude of choices can explain the lack of consistency surrounding ESG scores regarding disclosure standards and measurement methodologies. The plethora of different ESG scores has left investors frustrated and confused with their competing measurement methodologies. In fact, the latter vary so wildly that organisations have been able to cherry-pick the most appealing providers (Murray, 2021; Li and Polychronopoulos, 2020). It also makes it difficult to compare one ESG score methodology with another. As seen in (Berg et al., 2022), correlations between ESG scores are, on average, 0.54 and range from 0.38 to 0.71. Sustainalytics and Vigeo Eiris, both major ESG score providers, have the highest level of agreement with each other, with a correlation of 0.71. The correlations of the environmental dimension are slightly lower than the overall correlations, with an average of 0.53. This leads to capital markets to not adequately pricing the ultimate costs surrounding sustainable businesses. In addition, the lack of a uniform measurement approach can cause reputational damage, financial loss and regulatory fines. Overall, the major ESG score providers generally follow similar processes for calculating their scores. They use traditional research methodolo-

gies.[4] These include manually gathering publicly available information, sending surveys to companies, and receiving issuer feedback on the scores given to them. Thus, producing ESG scores appears to be manual, time-consuming, and prone to human bias and omissions. We take as our premise the view that there are two main barriers to producing accurate ESG scores: the limitations of the research methods used and the lack of robust data in the process. We seek to address this challenge by providing a free reproducible corpus with ESG relevance scores, and an ESG annotation methodology.

### 2.2. Automated Text Classification of Financial Texts and its relevance to ESG

Traditionally, quantitative financial data have been essential to understanding an investment's sustainability potential. In recent years, alternative data, such as news articles and social media, have become more important for assessing investment opportunities and financial market performance because they capture corporate information outside the realm of traditional financial data. Studies have shown that such information, especially news articles, affects the value and performance of organisations. This, in turn, has boosted research in financial news analysis (Hagenau et al., 2012; Kalyanaraman et al., 2014; Luss and d'Aspremont, 2008; Shah et al., 2018; Zhao et al., 2020). Even though research in AI-based ESG scores and trends is still in its infancy, it has been gaining interest. Several studies have used company disclosures and social media text to extract ESG information (Mehra et al., 2022; Raman et al., 2020; Nematzadeh et al., 2019; Shahi et al., 2014; Hisano et al., 2020). Other studies generated ESG scores or utilised ESG criteria to inform investment decisions and assess their impact on a company performance (Y.Aiba et al., 2019; Sokolov et al., 2021; Ribano and Bonne, 2010; Napier, 2019; Khan, 2019; Ghoul et al., 2011; Krueger et al., 2020; Guo et al., 2020; Brown, 2015).

While analysing company narrative and social media text carry relevant ESG information, we believe such approaches have limitations regarding data robustness and objectivity. To illustrate, detecting ESG relevance from corporate disclosure and earning call reports hampers ESG score objectivity and accuracy by excluding other important information sources, such as news articles. In addition, not all companies produce ESG or Corporate Social Responsibility (CSR) reports or include such sections in their annual reports. Companies tend not to disclose negative sentiments about themselves either voluntarily. Another limitation of the relevant studies is the lack of generalisation of their models to out-of-corpora models. This paper addresses these limitations by creating an ESG news article corpus that applies to other corpora domains.

### 2.3. Small data

Small data refers to an approach that requires less data but still offers useful insights. According to research by Gartner, 70% of organisations will shift their focus from big to small data by 2025 (Gartner, 2021). Small data seek to solve challenges stemming from scarce and disparate data, and historical data abruptly becoming obsolete and thus breaking AI models. To illustrate, breaking news can cause sudden changes in sentiment surrounding an organisation. In accordance with this, we argue that ESG-FTSE can be useful for obtaining relevant ESG insights. In line with recent work (Gururangan et al., 2020), we further demonstrate in our experiments that pretraining a model with a small corpus provides significant benefits: less computational resources and high accuracy in detecting ESG insights.

## 3. ESG-FTSE Corpus Development

This section describes the building, annotation process, and evaluation of the proposed corpus. Corpus development did not incur any costs and took ten days. Annotation took a week.

### 3.1. Approach Overview

The corpus construction process was divided into three phases: data collection, annotation, and evaluation. Each stage is described in detail in the following sections. In the first phase, we defined a set of criteria for data collection: category and data source. We selected News API as a retrieval method. We extracted news pieces about the top ten FTSE 100 Index companies to ensure suitability for financial market analysis. We extracted a total of 5,000 raw news articles. This was the initial, unlabelled version of the ESG-FTSE corpus. After this, we performed data cleaning on the initial corpus. In the second phase, we defined ESG relevancy criteria. We also introduced a three-level annotation schema. Due to the complexity of the annotation task, we defined a set of criteria for selecting annotators. Two annotators who met the requirements were selected. They were provided with training, clear guidelines and examples to minimise bias. We evaluated the ESG-FTSE corpus in the last phase and provided corpus statistics.

---

[4]Based on the lead author's experience in the financial sector and working with ESG score providers

## 3.2. Data Collection

### 3.2.1. Category definition and category selection

The Financial Times Stock Exchange 100 Index (FTSE 100 Index) is a share index tracking the 100 biggest companies by market capitalisation that are listed on the London Stock Exchange (LSEG), which is said to be the most used UK stock market indicator by investors.[5] For data extraction and annotation schema purposes, we define as category the name of each of the ten biggest FTSE 100 index constituents by market capitalisation.[6] [7] The company name "AstraZeneca" is an example of a category. The goal was to ensure the corpus is relevant for analysing the credentials of stock market companies.

### 3.2.2. ESG Topics

This paper considers a news article with ESG relevance as a topic. An ESG topic can include any news piece related to environmental, social, or governance matters. Table 1 lists some factors under each ESG pillar. We follow the SDG guidelines and the technical screening criteria under the EU taxonomy for sustainable activities. As per Annex I of the above-mentioned EU regulation, the taxonomy is a classification system that determines sustainability criteria for "economic activities aligned with a net zero goal and the broader environmental goals other than climate". For a detailed description of the screening criteria and scope of the regulation, please refer to the EU taxonomy.[8]

### 3.2.3. Data source selection

News API is a REST API that returns JSON results for current and historic news articles.[9] We utilise it for news article retrieval for each category. Being an established information retrieval method, it proved more suitable for the purposes of this research because it overcame the limitations associated with other news APIs and data collection techniques, such as RSS feed and web scraping. Namely, it allows the collection of historic articles effortlessly. It also has a wide range of endpoints, including full content. Furthermore, it facilitates reproducibility because the developer subscription is free. It ensures data robustness and non-bias as it returns results from over 80,000 news publications.

---

[5] https://www.londonstockexchange.com/
[6] As of 30 June 2021
[7] The top 11 constituents were taken due to Royal Dutch Shell being listed twice via different legal entities
[8] https://eur-lex.europa.eu/legal-content/EN/TXT/?uri=CELEX:32021R2139
[9] https://newsapi.org/

In addition, it solves data sparsity associated with ESG article collection.

### 3.2.4. News Article Extraction

The data extraction methodology is shown in Table 2. It is broken into three steps. The first step is collecting news articles for each category. The second step is corpus pre-processing. The third step is technical validation. In the first step, the same endpoints, language and time frame were used for each category. The aim was to achieve consistency. News articles were extracted in a csv format. The downloads were performed in ten batches over a ten-day period due to download limits associated with the free NewsAPI subscription. This produced ten csv files - one for each category. Two new columns, "Company Name" and "Number", were added to each file and filled with the corresponding category. For example, the column name of the AstraZeneca file was auto-completed with "AstraZeneca" in each row containing a news article. News articles in English were extracted between October 2018 and July 2021. To ensure data robustness, data were extracted by relevancy. A limit of 500 articles was set for each category. The following endpoints for each article were extracted via News API: title, author, source, description, content, publish date, and URL. In the data cleaning step, the following formatting changes were made to the data file to enhance understandability for future data use. Changes were made using Python. Duplicate news articles were removed. After examining the new corpus, more duplicates were noticed. Different news publications reusing the exact text caused some news pieces to be treated as unique by Python. Thus, another duplicate removal exercise was conducted. There is a word limit for csv files. To avoid losing article content and for consistency, a limit of 4,800 words per news article was applied. Last, the "content" endpoint was renamed to "Text". New columns were added: "Relevance Label" and "Primary Label". In the last step, news articles were checked for personally identifiable information, particularly e-mail addresses. This was done by searching for symbols and domains commonly used in e-mail addresses, i.e., "@" or ".com,". Author names were removed. For the purposes of this study, only the "Text", "Number", and "Label" columns were kept in the final corpus. All other columns were removed. After iterating over each category, 5,000 raw news articles were obtained. After removing duplicate articles, the final corpus consisted of 3,913 articles.

## 3.3. Annotation Process

Since this paper focuses on producing a corpus with ESG relevance labels, the paper deems rele-sustainable investing, economics and finance.

|  | **Factors** |
|---|---|
| Environ-mental | Greenhouse gas emissions, ground and air pollution, energy usage, carbon footprint, waste and water management, land use, biodiversity loss |
| Social | Labour practices, fair pay, equal employment opportunities, labour laws, workplace health and safety, responsible supply chain, community engagement, product quality, safety and access |
| Governance | Shareholder rights, board diversity, executive compensation, corporate governance, compliance, risk management, conflict of interest, corruption, accounting integrity |

Table 1: ESG pillars: key factors. The table is not exhaustive.

| **Data** | **Numerical or Text Value** |
|---|---|
| Category | AstraZeneca, Unilever, Diageo, HSBC, GlaxoSmithKline, Rio Tinto, BP, British American Tobacco, Royal Dutch Shell, BHP |
| News Articles | 500 per category |
| Time Period | 30/10/2018 – 31/07/2021 |
| Data Source | News API |
| Language | English |
| Raw End Points | title, author, source, description, content, publish date, URL |
| Final End Points | content - renamed to Text |
| Word Limit | 4,800 words per news article |
| Added Columns | Index, Company Name, Relevance Label, Primary Label |

Table 2: Data Collection methodology

| **Category** | **Text** | **ESG topic** |
|---|---|---|
| BP | BHP's oil exit would be better sooner than later | Environmental |
| BHP | Strike at BHP's Chile copper mines continue | Social |
| Rio Tinto | Rio Tinto appoints three women as non-executive directors | Governance |

Table 3: Examples of categories and ESG topics. The text column consists of news headlines.

### 3.3.2. Annotation Schema

Schemas from other domains and the SDGs inspired the annotation methodology.(Zampieri et al., 2019; Lee et al., 2022) Our methodology comprises a three-level schema: a binary classification (relevant versus irrelevant news articles), ESG classification (environmental, social, and governance-related news articles), and target company. Figure 1 outlines the annotation schema. Annotation was done manually. This method was adopted because it is considered the most precise method for document annotations. Two annotators who met the selection criteria were recruited. They conducted the labelling independently and according to two different levels of classification. The first layer is a binary classification: relevant versus irrelevant news articles. These are denoted by "1" and "0" respectively. Relevant news pieces must contain one category and at least one ESG topic. The second layer comprises an ESG classification: environmental, social, and governance-related news articles. These are represented in the "Primary Label" column as "E", "S", and "G", respectively. The ESG topic criteria are described in more detail in the ESG Topic section. It is to be noted that some articles may contain multiple ESG topics. We only classify the dominant topic, i.e., the primary topic. The third layer, the company name, was added during corpus pre-processing.

vant news articles that include both a category and at least one ESG topic. Table 3 shows examples of three ESG topics.

### 3.3.1. Annotator Selection

In the first step of the annotation process, we introduced a set of requirements for selecting annotators. This was necessary due to the multi-faceted nature of the annotation task. We decided to select experts recognised by industry and academia for their contribution to the ESG field. In addition, the specialists had to possess industry and academic experience in the following domains: ESG,

### 3.4. Annotation Evaluation

We computed inter-annotator agreement using Cohen's kappa (McHugh, 2012). We produced a Kappa score for both levels of the annotation schema. The Kappa scores show that high inter-annotator agreement was reached for both binary and ESG classification: 0.97 and 0.94, respectively. The small number of codes for each clas-sification task should be noted, as well as the high level of expertise of both annotators. We believe that it explains the substantial results.

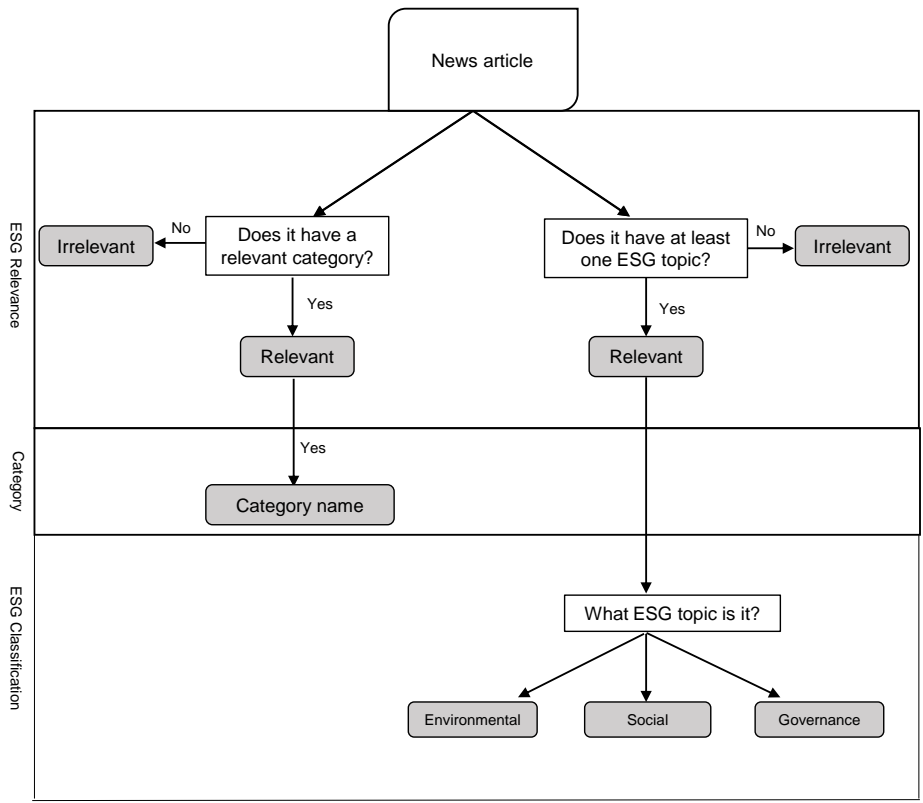

Figure 1: Annotation schema

## 4. ESG-FTSE corpus statistics

This section presents the ESG-FTSE corpus statistics. The corpus consists of 3,913 document-level annotations. First, a binary classification was performed. Each of the 3,913 news articles received either a "Relevant" or "Irrelevant" label: 1,178 and 2,735 news pieces, respectively. Next, the news articles with a "Relevant" label were also annotated according to an ESG-level classification. 418 news articles were classified as "Environmental", 218 news articles as "Social", and 542 news articles received a "Governance" label.

## 5. Experiments and Discussion

To validate the suitability of the proposed expert annotated corpus for ESG relevance detection, we implemented baseline experiments using both supervised and unsupervised learning methods for text classification. They showed that ESG-FTSE can successfully be used in different types of experiments to derive ESG insights from text. A second objective of the experiments was to describe the corpus and evaluate its quality by applying different tasks, representations, and machine-learning methods. A detailed description of the experiment and a discussion of the results are presented in the following sections.

### 5.1. Supervised Learning: Text Classification

We performed four supervised learning experiments: three for ESG relevance detection and one for ESG classification.

#### 5.1.1. ESG Relevance Detection

To detect ESG relevancy, we undertook three binary classification experiments. Due to the imbalanced nature of the corpus, a stratified K-Fold cross-validation was implemented in all experiments. It was essential to ensure the data were split randomly while maintaining the same class distribution in each subset. We determined that 5 splits with a class ratio of approximately 0.30 were most suitable. A different representation for each experiment was adopted to decide whether or not it would improve model performance. In general, SVM classifiers produce highly accurate results for binary classification problems (Schölkopf and Smola, 2018). Thus, an SVM classifier with a linear kernel and default parameters was chosen as a machine-learning method for all experiments. In addition, we used the pandas, sklearn, nltk and matplotlib packages in our experiments. Accuracy, precision, recall and F1 score with default parameters were adopted as evaluation metrics. Experiment 1 sliced the data in 67% train and 33% test. We used TF-IDF as representation (Sammut and Webb, 1970). Experiment 2 adopted a default training/validation split: 75% and 25% accordingly. In addition, we used *n-grams* and TF-IDF for feature extraction at different levels: word, n-grams (from bi- to four-grams), and character. Experiment 3 was an extension of Experiment 1, with *uni-grams* added to the pre-processing step. Table 4 presents the results of all experiments.

Table 4: ESG detection: binary classification. Results are rounded to two decimal places

| Experiment | Feature set | F1 Score | Accuracy | Precision | Recall |
|---|---|---|---|---|---|
| **Experiment 1** | TF-IDF | 79.09 | 88.62 | 86.33 | 72.96 |
| **Experiment 2** | TF-IDF, n-grams | 72.96 | 85.09 | 82.43 | 65.45 |
| **Experiment 3** | TF-IDF, uni-grams | 78.40 | 87.85 | 83.33 | 74.03 |

#### 5.1.2. ESG classification

In this experiment, we implemented a 5-fold stratified BERT model to classify Environmental, Social, and Governance labels. The smaller, pre-trained *bert-base-uncased* model was used. PyTorch, tqdm, BertTokenizer, pandas, sklearn and NumPy packages were adopted in the task. Data were split into 15% validation and 85% training sets. We used RandomSampler for training and SequentialSampler for validation. The training was conducted in 5 epochs. Due to class imbalance, weighted evaluation metrics were utilised: F1 score, accuracy, precision and recall. Results are displayed in Table 5

| Label | F1 | Accuracy | Precision | Recall |
|---|---|---|---|---|
| E | 0.92 | 0.86 | 1.00 | 0.86 |
| S | 0.65 | 0.48 | 1.00 | 0.49 |
| G | 0.60 | 0.43 | 1.00 | 0.43 |
| I | 0.97 | 0.95 | 1.00 | 0.95 |

Table 5: ESG classification results. Labels: E = Environmental, S = Social, G = Governance, I = Irrelevant. All scores are weighted average.

### 5.2. Unsupervised learning: Topic Modelling

Topic modelling is an unsupervised probabilistic algorithm that considers the problem of modelling discrete data, such as text corpora. The goal is to discover the main topics that occur in a set of documents by reducing their dimensionality. In our experiments, topic modelling was performed by building a latent Dirichlet allocation (LDA) model (Blei et al., 2003). Gensim, Nltk and Spacy packages were utilised. Hyperparameter tuning was performed in a series of hyperparameter sensitivity tests to improve the default LDA model's accuracy. The following hyperpatameters were tuned: *filter_extremes, random_state, update_every, chunksize, passes, alpha, eta and per_word topics.* The most optimal hyperparameter values were chosen based on the highest coherence score achieved on the LDA model. It achieved a coherence score of 0.60. The sensitivity tests are shown in Table 6. The most optimal hyperparameter values were chosen based on the highest coherence score achieved on the LDA model (Table 6, Test 3). Number of topics ($k$) is one of the most important LDA model inputs. Extracting the right number of topics largely depends on the dataset characteristics. Seven LDA models were built and compared by their coherence values to determine the most optimal $k$ number of topics. A limit of forty LDA models was set. The optimal number of topics was chosen based on the highest coherence value achieved on the final LDA model. *K=20* was selected because it achieved the highest coherence score of 0.60. More information is available in Appendix A. The evaluation was conducted via an intrinsic metric (coherence score C_v). The four most dominant topics are visualised via t-SNE in Figure 2. Entity salience was visualised via an interactive LDA model Intertopic Distance Map created via the pyLDAvis package. A snippet of it is available in Appendix B. In the interactive LDA model Intertopic Distance map, each bubble on the plot represents a topic. The larger the bubble, the more prevalent the topic. Additionally, a bar chart representing the top 30 most salient keywords that form a selected topic is available in the interactive LDA model Intertopic Dis-tance Map. It reveals high ESG salience in each topic.

Table 6: LDA model sensitivity tests

| | Hyperparameters | | | | |
|---|---|---|---|---|---|
| Test | Alpha | Beta | filter_extremes | Perplexity | Coherence |
| 1 | 0.01 | 0.9 | no_below=10 no_above=0.20 | -7.597 | 0.538 |
| 2 | 0.01 | 0.5 | no_below=10 no_above=0.15 | -7.670 | 0.565 |
| 3 | 0.03 | 0.5 | no_below=10 no_above=0.15 | -7.652 | 0.597 |
| 4 | 0.05 | 0.5 | no_below=10 no_above=0.15 | -7.673 | 0.552 |
| 5 | 0.04 | 0.5 | no_below=10 no_above=0.15 | -7.665 | 0.572 |
| 6 | 0.04 | 0.4 | no_below=10 no_above=0.15 | -7.668 | 0.551 |

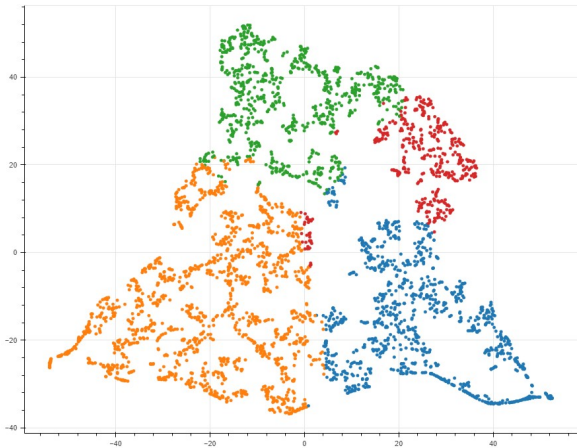

Figure 2: t-SNE clustering of the top 4 topics

### 5.3. Discussion

We evaluated two different classification tasks: ESG relevance detection and ESG classification. Results demonstrated that the ESG-FTSE corpus is highly quality and can be used successfully for ESG knowledge extraction in supervised and unsupervised models. In addition, the experiments proved that valuable ESG insights can be obtained even from low-volume data. Last, the topic modelling experiment provided a thorough context and description of the corpus. We implemented three baseline experiments for ESG relevance detection. All of them achieved high performance. The best model for this task, Experiment 1, obtained a 79% F1 score. For ESG detection, we performed a stratified 5-fold BERT experiment. As shown Table 5, high performance was also achieved for this task. To illustrate, adequate F1 scores were obtained for each label. We take at our premise the view that this suggests a balanced model. The unsupervised experiment also produced high results. Probabilistic topic models like LDA always produce topic outputs. However, making them valuable and meaningful for this research demanded capturing the correct information, i.e., the minority ESG class. Despite the class imbalance of the corpus, the topic modelling experiment successfully extracted relevant ESG information. The t-SNE plot of the four most dominant topics indicates that the most similar documents are grouped in well-defined clusters (Figure 2). The most salient words also demonstrate the robustness of the model for a given topic. The results validated our initial view that a well-defined annotation schema yields a good model performance in complex and subjective domains like ESG.

### 6. Conclusion

The most significant contribution of this study is presenting a free, reproducible corpus to facilitate data sharing in a standardised framework. Being the first corpus with ESG relevance, ESG-FTSE provides the first-of-its-kind annotation schema. In addition, it offers a novel solution to the bias associated with ESG scores. The present study revealed class imbalance due to data sparsity. Instead of trimming the corpus or boosting the minority class, we trained on all possible instances to maximise coverage. The evidence confirmed that small data can be insightful in obtaining relevant ESG insights.

# 7. Ethical Considerations and Limitations

A developer News API license was obtained to download the news articles. Data were downloaded and used following News API Terms. [10] According to the provider, all data are publicly available. No personal data, such as user analytics or cookies, were used in this study. News API is compliant with UK and EU data laws and directives. To illustrate, its privacy policy states that News API "has been prepared to fulfil the obligations under Art. 10 of EC Directive n. 95/46/EC, and under the provisions of Directive 2002/58/EC, as revised by Directive 2009/136/EC, on the subject of Cookies." [11]. Thus, all data used for this study is considered ethical and lawful.

# 8. Bibliographical References

---

[10] https://newsapi.org/terms
[11] https://newsapi.org/privacy

# 9. Appendix

## A. Topic Modelling

```
1  # Show graph
2  limit=40; start=2; step=6;
3  x = range(start, limit, step)
4  plt.plot(x, coherence_values)
5  plt.xlabel("Num Topics")
6  plt.ylabel("Coherence score")
7  plt.legend(("coherence_values"), loc='best')
8  plt.show()
```

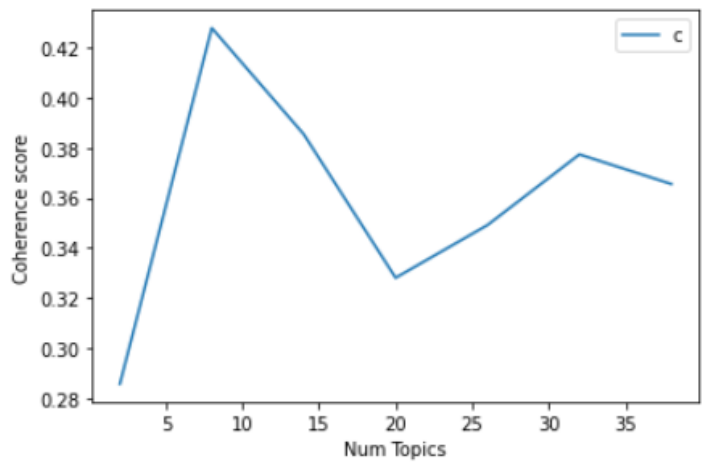

Figure 3: Number of topics.

## B. pyLDAvis visualisation tool

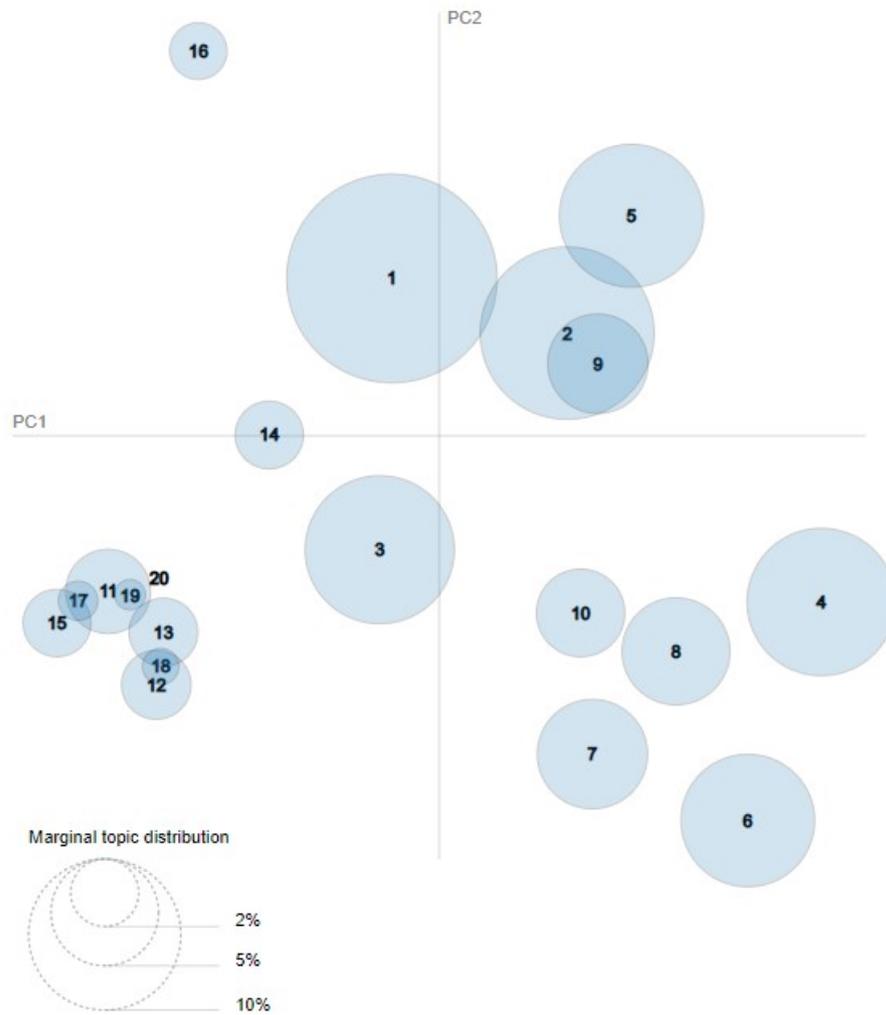

Figure 4: Intertopic Distance Map (via multidimensional scaling). Each bubble on the plot represents a topic. The larger the bubble, the more prevalent the topic.

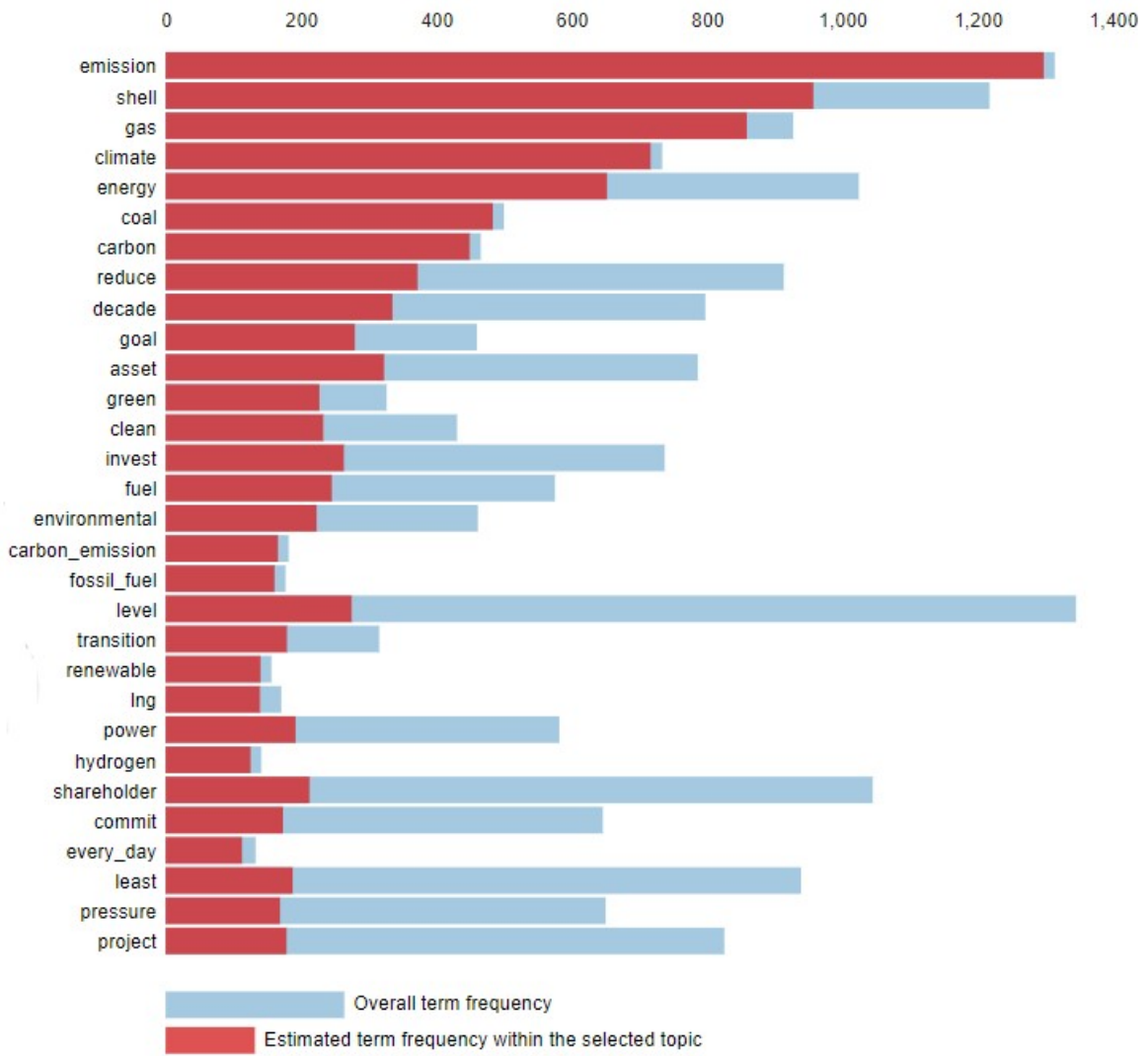

Figure 5: Salient words in an example topic. It has high ESG salience.